\documentclass[10pt,a4paper]{article}

\usepackage[margin=2.4cm]{geometry}
\usepackage[T1]{fontenc}
\usepackage[utf8]{inputenc}
\usepackage{cite}
\usepackage{amsmath,amssymb,amsfonts}
\usepackage{bm}
\usepackage{algorithmic}
\usepackage{graphicx}
\usepackage{textcomp}
\usepackage{booktabs}
\usepackage{tabularx}
\usepackage{pifont}
\usepackage{caption}
\usepackage{authblk}
\usepackage[hidelinks]{hyperref}

\newcommand{\preprintnotice}{%
This work has been submitted to \emph{IEEE Access} for possible
publication. Copyright is held by IEEE. Personal use of this material
is permitted; permission from IEEE must be obtained for all other
uses, in any current or future media, including reprinting or
republishing this material for advertising or promotional purposes,
creating new collective works, for resale or redistribution to servers
or lists, or reuse of any copyrighted component of this work in other
works. This version may no longer be accessible without notice.
}

\title{\bfseries A Controlled Comparison of Manual and Teleoperated
Intraocular Instrument Motion for an Input Device}

\author[1,*]{Korab Hoxha}
\author[2,*]{Mirza Imamovic}
\author[1]{Angelo Henriques}
\author[1,3]{M. Ali Nasseri}

\affil[1]{Department of Ophthalmology, School of Medicine and Health,
TUM University Hospital, Munich, Germany}
\affil[2]{Chair of Ergonomics, Munich Institute of Robotics and
Machine Intelligence, Technical University of Munich, Munich, Germany}
\affil[3]{Department of Biomedical Engineering, University of Alberta,
Canada}
\affil[ ]{\vspace{0.3em}\normalsize
Corresponding author: Korab Hoxha (e-mail:
\texttt{korab.hoxha@tum.de}).\\
$^{*}$The first two authors contributed equally to this work.\\
This work is partially supported by NSK Ltd.}

\date{}

\begin{document}
\maketitle
\thispagestyle{plain}

\renewcommand{\thefootnote}{}
\footnotetext{\small\preprintnotice}
\renewcommand{\thefootnote}{\arabic{footnote}}
\setcounter{footnote}{0}

\begin{abstract}
\noindent
Input devices for robotic microsurgery are frequently described as preserving the surgeon's
trained technique, but the claim is rarely measured. We compared manual and teleoperated intraocular
instrument motion with the trocar constraint, the instrument, the eye model and the tracking source common to both conditions, so that the control interface was the only factor varied. Prior comparisons cannot hold the instrument fixed, because a robotic instrument is not the tool used manually. Sixteen participants performed a navigation task on a commercial ophthalmic simulator by hand and through a three-degree-of-freedom input device commanding a five-joint robot. Task outcome was equal but at ceiling: every participant acquired all five targets under both interfaces with no retinal or lens injury. Execution differed on every measure. Teleoperated trials took three times as long at a quarter of the median speed, covered less than half the angular working range, and were broken into 3.5 times as many separate movements. Completion time and movement fragmentation improved substantially across four trials of practice and had not plateaued; the measures set by the configured rate ceiling and joint limit changed the least. Finger activity doubled and pinch variability tripled, so reducing instrument degrees of freedom redistributed manual effort rather than reducing it. The interface preserves the outcome and reshapes the execution.
\end{abstract}

\vspace{0.5em}
\vspace{1em}

\section{Introduction}
\label{sec:intro}
Vitreoretinal surgery treats disorders of the retina and vitreous
through instruments introduced into the eye via trocars in the sclera. The relevant anatomy is small and fragile: the internal limiting membrane is 2--3\,$\mu$m thick, retinal veins are
of order 100\,$\mu$m in diameter, and the retina does not regenerate, so
inadvertent contact can cause permanent visual loss. Subretinal injection, an
emerging route for gene and cell therapy, requires placing a needle tip within
tens of micrometers of the retinal surface and holding it there.

\begin{figure}[!t]
\centering
\includegraphics[width=0.42\linewidth]{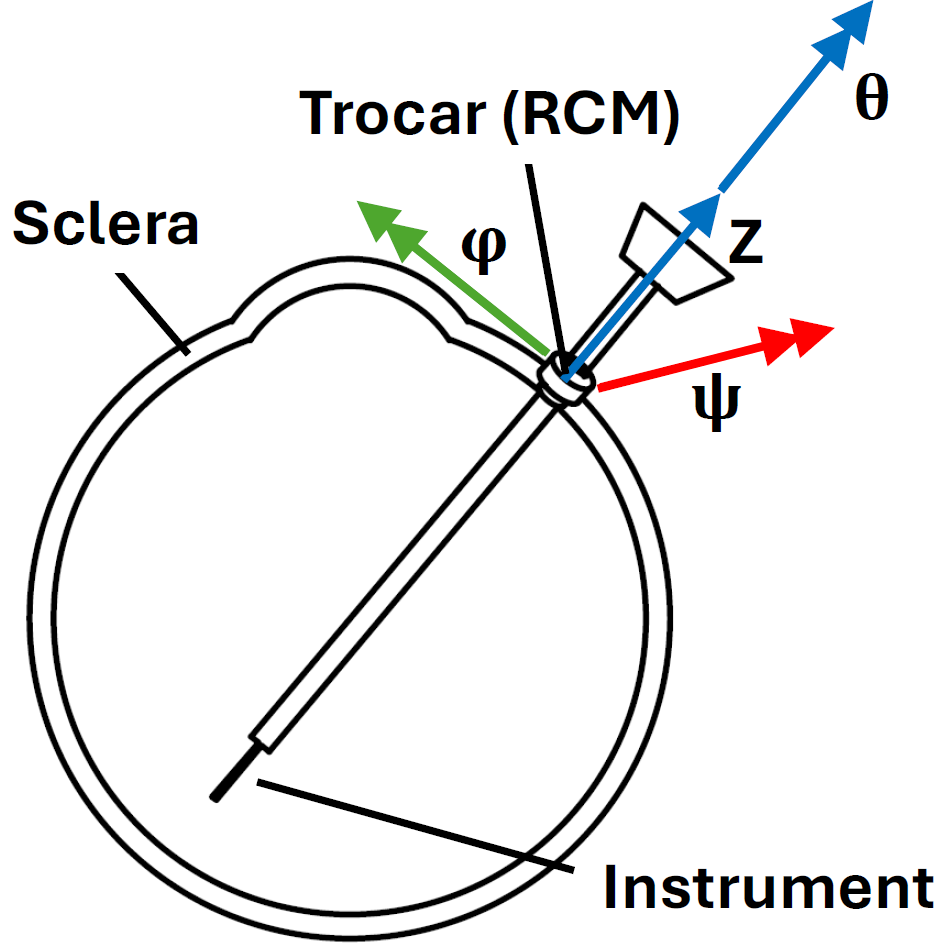}
\caption{Cross-section of an eye with an instrument inserted through a trocar.
The degrees of freedom ($\theta, \phi, \psi, z$) of the tool are given by the arrows}
\label{fig:eye}
\end{figure}

Three features of the procedure shape any robotic assistance. The trocar imposes
a remote center of motion (RCM) at the sclera, so once inside the eye the
instrument retains four degrees of freedom: three rotations about the entry point
and one translation along the instrument axis (Fig.~\ref{fig:eye}) \cite{b2}.
Two of the rotations aim the instrument tip across the retina; the third is a roll about the
instrument axis, which matters for tools with a directional tip but not for the
positioning task studied here. The eye is a closed, fluid-filled cavity, so
lateral force at the entry port stresses the sclera and can displace the globe.
And the surgeon works under a microscope with limited depth cues, at
magnifications where hand tremor is visible in the field.

That last point sets the precision limit. Structures of interest are tens of
micrometers across, while physiological hand tremor during vitreoretinal surgery
spans 100--200\,$\mu$m peak-to-peak \cite{b1}. Robotic assistance has been
pursued on the premise that scaling motion and filtering
tremor can extend the surgeon's reach beyond that limit.

The effectiveness of such a system depends on the interface through which the
surgeon commands it. A robot that can position an instrument to a few
micrometers is only as useful as the operator's ability to specify where those
micrometers should be. This has motivated a class of application-specific
input devices for ophthalmic surgery, and a recurring design argument: that an
input device should preserve the movements the surgeon has already spent years
learning, rather than require new ones.

That argument is widely made and rarely tested. Input devices are
characterized by specification: degrees of freedom (DOF), workspace, peak
force, update rate. They are evaluated by task outcome: completion time, error
counts, subjective workload. A device may be described as intuitive, or as
preserving natural surgical technique, without either property being measured.
What would make the claim testable is a direct comparison of the motion a
surgeon produces through the interface against the motion the same surgeon
produces by hand, on the same task.

Such comparisons exist for laparoscopic and general surgery, where the dynamics
of the input device have been shown to alter operator movement measurably
\cite{b11,b12,b13,b14,b15}. In every case the instrument at the tissue differs
between conditions: a wristed robotic instrument is not the tool the surgeon
holds manually. Where the manual condition is open technique rather than
laparoscopic, the kinematic constraint differs as well. In intraocular surgery
neither confound is necessary. The trocar imposes a remote center of motion
(RCM) in both conditions, and a robot can hold the same unmodified instrument
the surgeon would hold, tracked by the same system. The comparison can therefore
be made  while holding the physical instrument, trocar constraint, eye model, and measurement source constant, allowing the effects of the teleoperation pathway and its control architecture to be examined without simultaneously changing the surgical tool or environment.

This paper reports such a comparison for an input device we call EMIRAS (Ergonomic Manipulator for Intraocular Robotic-Assisted Surgery), built explicitly on the manual-similarity premise. EMIRAS is held in a pen grasp with the wrist
rested, pivots about a fulcrum below the fingers, and multiplexes three sensed
inputs onto the five joints of a robot, with aiming mapped position-to-position
in a frame aligned to the microscope view. Sixteen participants (three
vitreoretinal surgeons, six residents, seven novices) performed a navigation
task on a commercial ophthalmic simulator, manually and through the device, with
the same instrument, the same eye model, the same trocar port and the same
tracking source in both conditions.

The purpose of this comparison is not simply to determine whether the
teleoperated interface can complete the same task as manual operation, but to
determine whether it elicits similar movement behavior. We therefore examine
task success together with movement speed, path length, angular workspace,
movement fragmentation, mode-engagement behavior, practice effects, and
finger-level kinematics. This allows us to identify which characteristics of
manual microsurgical movement are retained and which are transformed by the
teleoperation architecture.

The contributions of this work are:

\begin{enumerate}
\item \textbf{A complete system description} of a three-DOF input device controlling a
five-DOF intraocular robot through phase-specific mode multiplexing, with all
control parameters, sensing resolutions, timing and workspace limits reported as
configured or measured rather than as specified.

\item \textbf{A modality-agnostic basis for comparing intraocular instrument
motion.} The RCM is recovered per trial from the instrument trajectory by
least-squares line intersection, the standard pivot-calibration formulation, and
motion is decomposed into insertion depth and aiming angles
about it. The method is not new; applying it per trial as a common frame in
which manual and teleoperated motion become directly comparable, despite a
depth-dependent displacement scaling, is what this contributes.

\item \textbf{A controlled comparison} in which the physical instrument, eye model,
trocar constraint, task, and measurement chain are common to both conditions,
allowing the effects of the teleoperation pathway to be examined without
confounding them with changes in the surgical tool or environment.

\item \textbf{Quantitative characterization of effects the literature does not
report}: clutch duty cycle and engagement statistics for a surgical input
device, tremor amplitude projected in a common physical unit from the handle through the robot joints to the instrument tip, and the learning
trajectory of a novel interface against a familiar one. Finger-level tremor is
reported separately and in relative terms only, since it does not convert to
the same unit.
\end{enumerate}

Section~\ref{sec:related} reviews input devices for microsurgery and the
existing evidence on manual versus teleoperated movement.
Section~\ref{sec:system} describes the system. Section~\ref{sec:methods} details
the experimental design and the analysis. Section~\ref{sec:results} reports the
results, Section~\ref{sec:discussion} discusses their implications for interface
design, and Section~\ref{sec:conclusion} concludes.

\section{Related Work}
\label{sec:related}

\subsection{Input Devices for Robotic Microsurgery}
Input devices for surgical robotics fall broadly into general-purpose
commercial devices and application-specific designs.

The commercial category is dominated by parallel-kinematic interfaces. The
Force Dimension omega, delta, sigma and lambda families are the reference
designs \cite{b4,b5}: the omega.7 provides seven active DOF with full gravity
compensation and up to 8\,N of grasping force; the sigma.7 was the first
commercial interface to offer seven active DOF including force-feedback
grasping, and was adopted as the console interface for the MiroSurge bimanual
system \cite{b6}; the lambda.7 is designed explicitly for integration into
surgical consoles. These devices have reached clinical products: omega.3 in the Medrobotics Flex system and lambda in the Titan Medical SPORT platform.
Serial-chain interfaces such as the 3D Systems Touch offer larger workspaces at
the cost of achievable feedback force, a trade-off examined comparatively by
Zareinia \emph{et al.} \cite{b7}.

Application-specific input devices for vitreoretinal surgery are fewer. The
Preceyes Surgical System \cite{b8} and the interface of Gijbels \emph{et al.}
\cite{b9} are both four-DOF controllers matched to the intraocular workspace,
incorporating motion scaling and tremor filtering. Takahashi \emph{et al.}
\cite{b10} take a different approach, providing six DOF but constraining the
operator's arm posture so that only minimal hand motion is required.

What unites these designs, general-purpose and specialized alike, is that they
are characterized by specification (degrees of freedom, workspace volume, peak force, update rate) rather than by the correspondence between the
motion they elicit and the motion the surgeon already performs.

\subsection{Kinematic Comparison of Manual and Teleoperated Operation}
The most directly relevant body of work is a sustained program by Nisky,
Okamura and colleagues comparing freehand movement against da Vinci
teleoperation. They report target-acquisition error, movement speed and smoothness
\cite{b11,b12}; a nine-metric comparison across six robotic surgeons and ten
novices \cite{b13}; arm joint-angle variability \cite{b14}; and trial time, path
length, coordination and learning in teleoperated versus open needle driving
\cite{b15}. Their central finding, that the dynamics of the input device measurably alter
operator movement, motivates the present work.

Comparable comparisons exist in laparoscopy. Balasubramani \emph{et al.}
\cite{b16} measured hand movement counts and durations across open,
laparoscopic and robotic technique using electromagnetic tracking; Berguer and
Smith \cite{b17} compared arm displacement and muscle activation between robotic
and laparoscopic technique; Lee \emph{et al.} \cite{b18} compared manual
endoscopic against telerobotic simulation on duration and upper-extremity
posture; and Zihni \emph{et al.} \cite{b19} compared completion time and errors
across platforms.

Two features of this literature define the gap addressed here. First, the
instrument differs between conditions: a wristed robotic instrument and a
handheld laparoscopic instrument are not the same physical object, so tool-side
differences are confounded with interface differences. Laparoscopic manual
technique is itself trocar-constrained, so the kinematic constraint is shared
in those comparisons, but the tool is not; where the manual condition is open
technique the constraint differs too. Second, the domain is laparoscopic or
general surgery. We are not aware of an equivalent comparison for intraocular
microsurgery, where motion scaling is an order of magnitude higher and the
relevant amplitudes approach the tremor floor.

\subsection{Workspace, Scaling and Clutching}
Workspace has been analyzed extensively as a property of instruments and
manipulators \cite{b20,b21,b22,b23}. Effects on the input-device side have
received less attention: Kim \emph{et al.} \cite{b24} examined misalignment
between input device and instrument, and Maddahi \emph{et al.} \cite{b25}
proposed kinematic indices of the input device as performance measures.

Clutching, which is releasing and repositioning the input device to extend
effective range, is near-universal in teleoperation, and indexing strategies and
workspace-drift control for small-workspace haptic devices have been studied
\cite{b41}, as has clutch usage in clinical robotic systems \cite{b42}. What we
have not found is a treatment of engagement behavior as a measurable
consequence of a specific control design: duty cycle, engagement duration and
engagement count reported for a surgical input device alongside the workspace
and mode
structure that produce them.
That is the gap this work addresses. As Section~\ref{sec:discussion} reports,
the outcome was not the one the geometry alone would predict.

\subsection{Tremor Measurement}
Physiological tremor in vitreoretinal surgery was characterized by Singh and
Riviere \cite{b1} using an instrumented tool with six-DOF inertial sensing,
establishing the 100--200\,$\mu$m amplitude figure the field still cites.
Subsequent work measures tremor at the instrument, by stereo vision
\cite{b26}, optical coherence tomography \cite{b27}, position and force sensing
\cite{b28}, and computer-vision tracking from microscope video \cite{b29,b30}.
Recent ophthalmic work quantifies suppression by passive support robots using
instrument-mounted accelerometers \cite{b31,b32}, and Verrelli \emph{et al.}
\cite{b33} measured how tremor propagates during microsurgery.

Across this literature tremor is measured at a single point: at the instrument,
or at a handheld tool. Measurement at more than one point in a teleoperation chain, at the input device
and projected to the instrument tip in a common unit with the hand measured
separately in relative terms, we have not found reported; and Section~\ref{sec:results} shows why the distinction matters, since
the answer differs qualitatively depending on where in the chain one measures.

\subsection{Hand-Level Measurement in Surgery}
Instrumented gloves have been used chiefly for skill assessment, in open surgery
\cite{b34,b35}, neurosurgical simulation \cite{b36}, and laparoscopic training
\cite{b37,b38,b39}. Ophthalmic application is recent and rare. Akada \emph{et al.} \cite{b40}
assessed surgical skill in simulated intrascleral intraocular lens fixation
using stretchable strain sensors on all fingers. To our knowledge this is the
only prior finger-level kinematic measurement in ophthalmic microsurgery, and it
is confined to manual technique.

\subsection{Position of This Work}
This work characterizes an input device by the correspondence between the motion
it elicits and the motion of manual technique, measured at the hand and at the
instrument. It also reports measurements the literature above does not: clutch
duty cycle and engagement statistics for a surgical input device, and tremor
projected in a common unit from the handle to the instrument tip.

\section{System Description}
\label{sec:system}

\subsection{Design Rationale}
EMIRAS follows the premise set out in Section~\ref{sec:intro}. The stylus is
held in a pen grasp with the wrist rested, matching the posture used with a
manual instrument, and pivots about a point at its distal end, so the
hand rotates about a fulcrum below the fingers, the same relationship the hand
has to an instrument constrained by a trocar.

The device multiplexes three sensed
inputs onto five robot joints through phase-specific modes rather than mapping
every axis simultaneously.

\subsection{Mechanical Design and Instrumentation}
\begin{figure}[!t]
\centering
\includegraphics[width=0.85\linewidth]{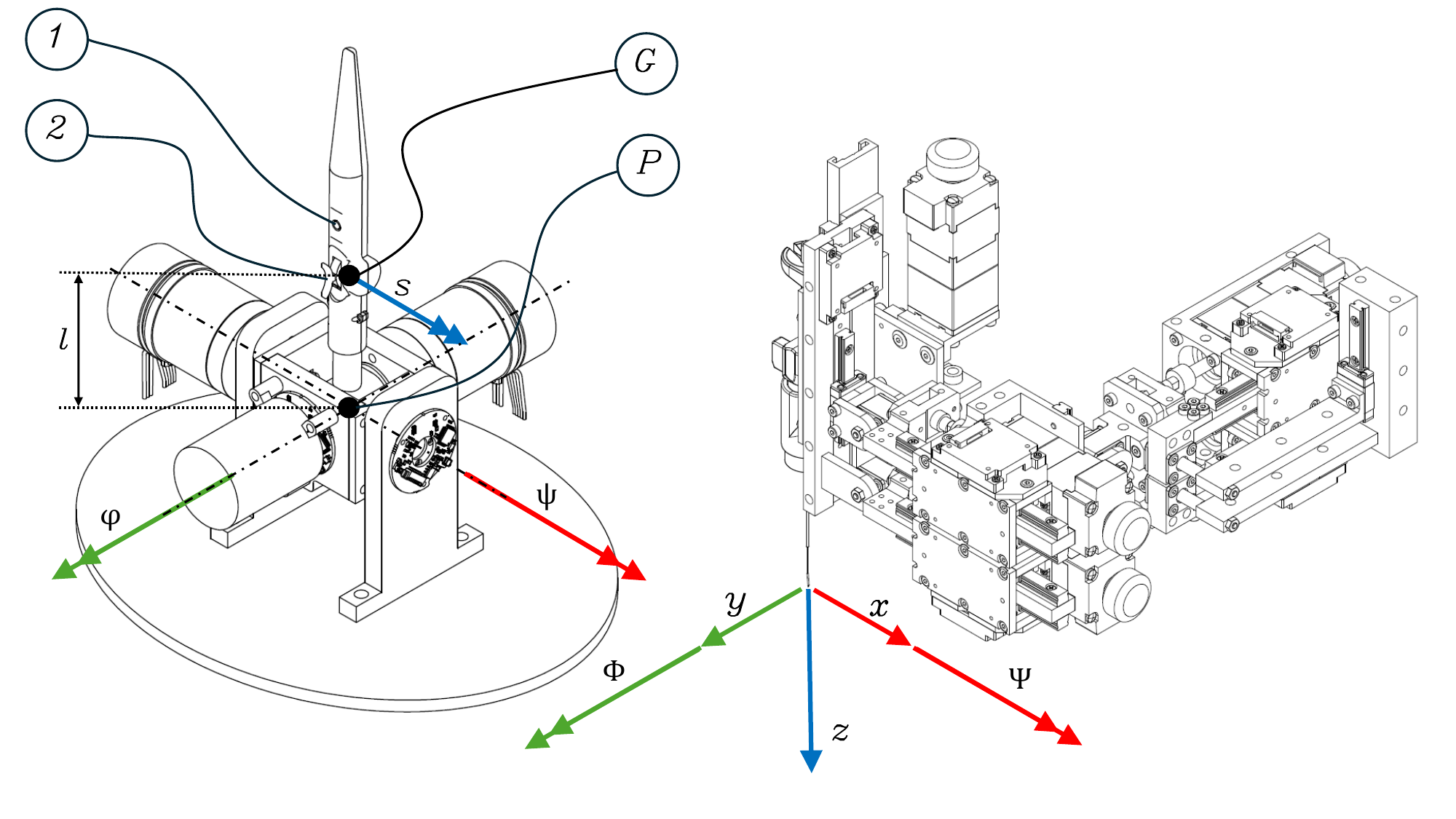}
\caption{Input device (left) and robot (right), with the correspondence between commanded motions. On the device, the two gimbal rotations $\psi$ and $\varphi$ act about orthogonal axes intersecting at the pivot point $P$; the operator grips the stylus at $G$, a distance
$l = 100$\,mm from $P$. The momentary button \ding{172} selects between aiming and insertion, and the spring-loaded lever \ding{173} contributes the third input $s$. On the robot,
$\varPsi$ and $\varPhi$ are the corresponding rotations of the instrument about the
trocar.}
\label{fig:device}
\end{figure}

A stylus is mounted on a two-axis gimbal whose axes intersect at a pivot point
$P$ at the stylus's distal end, giving rotations $\psi$ and $\varphi$ (Fig.~\ref{fig:device}). The
operator grips the stylus at $G$, $l = 100$\,mm from $P$. A spring-loaded lever
on the stylus returns to center when released; its angular deflection $s$ is the
third input and is kinematically independent of the gimbal. A momentary button
on the stylus selects between rotation and insertion.

Positions are transmitted unfiltered; the only low-pass filter in the firmware
acts on an internally derived velocity estimate used by the force-rendering law,
which was inactive during this study. Table~\ref{tab:spec} lists the sensing, actuation, timing and range parameters.

\begin{table}[!t]
\caption{\textbf{System Specification}}
\label{tab:spec}
\renewcommand{\arraystretch}{1.25}
\centering
\small
\begin{tabularx}{0.8\linewidth}{@{}lX@{}}
\toprule
Item & Specification \\
\midrule
\multicolumn{2}{@{}l}{\textit{Sensing}}\\
Gimbal encoders ($\psi$, $\varphi$) & RLS AksIM-2, 17-bit absolute, SPI \\
Angular resolution & $360^{\circ}/131072 = 0.00275^{\circ}$ \\
Linear resolution at $G$ & 4.8\,$\mu$m \\
Lever encoder & RLS RM08, 12-bit absolute \\
Lever resolution & $360^{\circ}/4096 = 0.088^{\circ}$ \\
\addlinespace
\multicolumn{2}{@{}l}{\textit{Actuation}}\\
Actuators  & $2\times$ maxon EC45 flat, 80\,W \\
Gearing & maxon GP42C planetary, single stage \\
Drives & maxon EPOS4 Compact 50/5, CAN \\
Gravity compensation & Passive counterweight \\
Vibrotactile unit & Stylus-mounted, PC-scheduled \\
\addlinespace
\multicolumn{2}{@{}l}{\textit{Electronics and timing}}\\
Microcontroller & STM32 F767ZI \\
Sensor sampling & 1\,kHz coherent snapshot \\
Uplink rate to host & $\sim$490\,Hz measured \\
Host control loop & 100\,Hz \\
Robot control period & 10\,ms \\
Button debounce & 5\,ms \\
\addlinespace
\multicolumn{2}{@{}l}{\textit{Ranges}}\\
Gimbal range ($\psi$, $\varphi$) & $\pm 40^{\circ}$ \\
Lever range ($s$) & $\pm 30^{\circ}$ mech., $\pm 22^{\circ}$ mapped \\
\bottomrule
\end{tabularx}
\end{table}

\subsection{Input Variables and Scaling Geometry}
The input vector is
\begin{equation}
\mathbf{q}_{M} = [\psi,\; \varphi,\; s]^{\mathrm{T}}
\end{equation}
where $\psi$ and $\varphi$ are the two gimbal rotations about orthogonal axes
intersecting at $P$, and $s$ is the lever deflection. Both gimbal angles are
measured directly by absolute encoders; no forward-kinematic model is
interposed, because the control path maps input angles to robot joint angles
without passing through Cartesian grip position. Because $s$ acts on a mechanism
carried by the stylus rather than on the gimbal, it does not displace the grip
point, which is the mechanical basis for separating aiming from insertion.

The one geometric relation the mapping does use is the displacement ratio
between grip and tool tip. The grip traverses an arc of radius $l = 100$\,mm
about $P$, while the instrument tip traverses an arc of radius $L$ about the
trocar (Fig.~\ref{fig:device}). With the angular scaling $g$ defined below,
\begin{equation}
\frac{\Delta p_{\mathrm{grip}}}{\Delta p_{\mathrm{tip}}} = \frac{g\,l}{L}.
\end{equation}
$L$ varies continuously as the instrument advances, so the operator's effective
hand-to-tip displacement gain is depth-dependent within a single insertion.

\subsection{Robot}
The robot is a parallel--serial manipulator developed and previously
characterized by Nasseri \emph{et al.} \cite{b3}. Its mechanism reduces kinematically to a
serial chain of three translational joints ($d_1$, $d_3$, $d_5$) and two
rotational joints ($\theta_2$, $\theta_4$), whose combined effect is the instrument's two aiming rotations $\varPsi$ and $\varPhi$ about the trocar (Fig.~\ref{fig:device}), and that reduced chain is the model
used throughout this work; the full mechanism and its derivation are given in
the cited work and are not repeated here.

The rotational joints have a mechanical range of $\pm 9^{\circ}$, further
restricted to $\pm 8^{\circ}$ by a software excursion limit. Under the RCM
constraint $d_1$ and $d_3$ follow from the rotations and the fixed trocar
position; they are commanded independently only in Free mode
(Section~\ref{subsec:mapping}).

The robot holds an unmodified Eyesi Surgical instrument, so the instrument at
the tissue interface is identical in the manual and teleoperated conditions.

\subsection{Input-Device-to-Robot Mapping}
\label{subsec:mapping}
\subsubsection{Frame alignment}
The robot's rotational joints are not parallel to the microscope view presented
to the operator. Device angles are therefore rotated into the robot frame by a
fixed $\theta = 300^{\circ}$ before any control law acts:
\begin{equation}
u_2 = \frac{\cos\theta\,\varphi - \sin\theta\,\psi}{g},\qquad
u_4 = \frac{\sin\theta\,\varphi + \cos\theta\,\psi}{g}.
\end{equation}
This aligns commanded motion with the visually perceived motion, preserving the
correspondence between hand direction and visual direction that the operator has
under a microscope in manual surgery. Both handle axes contribute to both robot
axes.

\subsubsection{Motion scaling}
The scaling factor g in (3) sets the ratio between commanded handle angle and commanded robot angle. We fixed $g = 3.0$, so that the robot's full $\pm 8^{\circ}$ rotational range is reached at $\pm 24^{\circ}$ of gimbal deflection. The value of g follows the premise in Section I that motion scaling extends manual precision beyond the tremor floor: a 3:1 reduction maps hand-scale excursions onto the sub-tremor-amplitude motion the retina requires, without demanding gimbal precision beyond what the encoders and the operator's hand can reliably produce.

\begin{figure}[!t]
\centering
\includegraphics[width=0.92\linewidth]{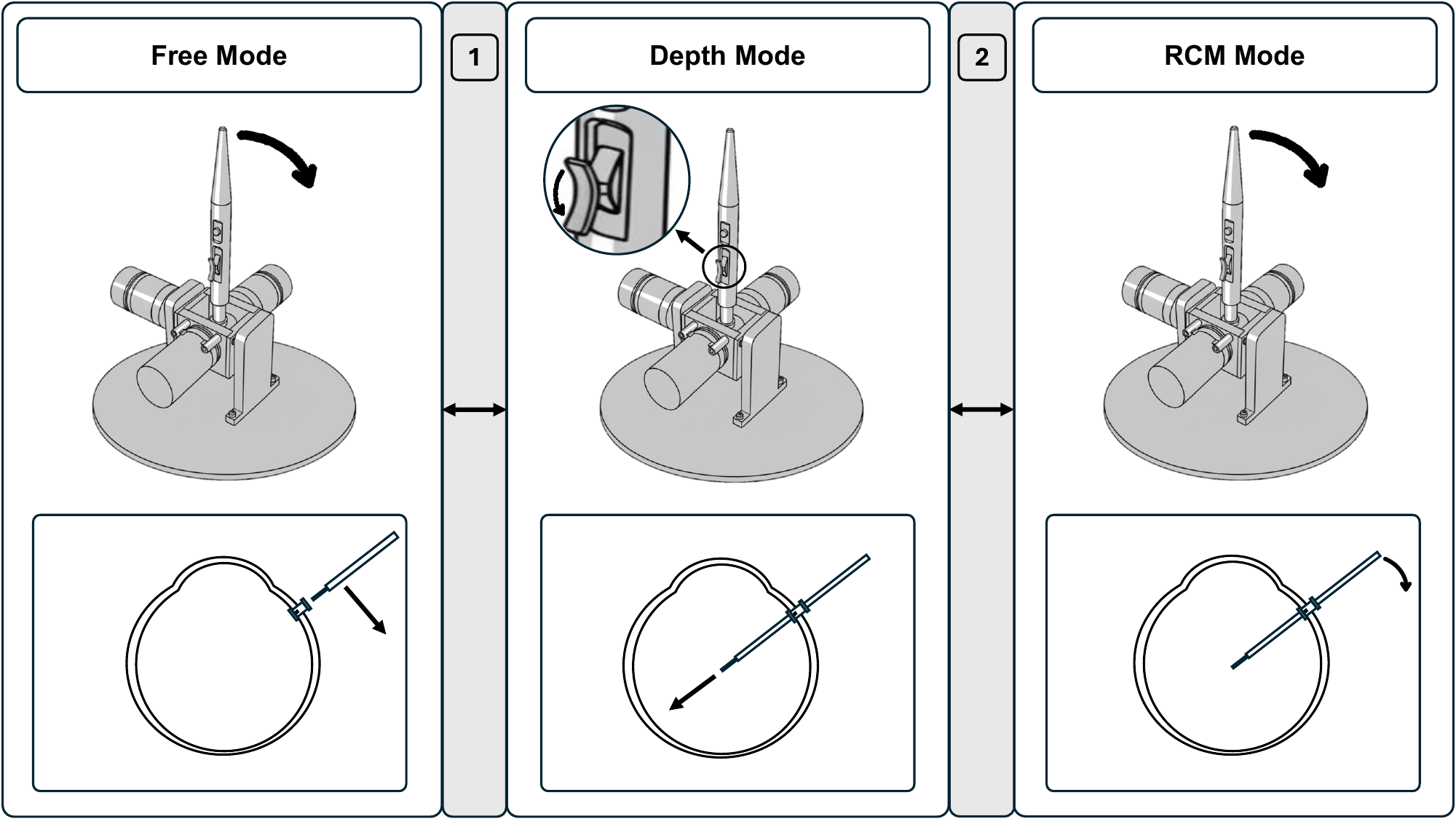}
\caption{The three operating modes and the instrument state each produces. Free
mode positions the instrument outside the eye and is used only during setup.
Depth mode is the default once inside: the rotational degrees of freedom are
locked and the slider drives insertion along the instrument axis. RCM mode,
active while the movement button is held, aims the instrument about the trocar.
The transition between the extraocular and intraocular configurations is made in
software on the host; the movement button switches between Depth and RCM modes.}
\label{fig:modes}
\end{figure}

\subsubsection{Operating modes}
The device has three modes, corresponding to the phases of the procedure
(Fig.~\ref{fig:modes}).

In Free mode, used before the instrument enters the eye, the two gimbal
rotations and the lever command Cartesian velocities of the robot's three
translational joints, with the rotational joints held. This positions the
instrument at the trocar during setup. Because the RCM constraint does not yet
apply, this is the only mode in which $d_1$ and $d_3$ are commanded
independently. It is used during setup and is not exercised in any recorded
trial.

Once the instrument is inside the eye, the button multiplexes the two
intraocular modes. Pressed selects Aiming or RCM mode: the gimbal rotations drive the
robot's rotational joints under the RCM constraint, with $d_5$ held. Released
selects Insertion or Depth mode: the lever drives $d_5$ while the rotational joints hold the last commanded pose. The two are mutually exclusive by construction, so
commanding insertion requires releasing the aiming command. This is not a
conventional clutch: the handle-to-joint mapping is absolute, so releasing and
repositioning the handle does not extend the robot's reach.

\subsubsection{Rotational law}
A proportional velocity command on the tracking error
$e = u - \theta_{\mathrm{robot}}$:
\begin{equation}
v_{\mathrm{cmd}} = G\,e,\qquad
\mathrm{offset} = \mathrm{clamp}\!\left(\pm 100\,\frac{v_{\mathrm{cmd}}}{V}\right)
\end{equation}
with $G = 4.0$\,s$^{-1}$, $V = 1.5^{\circ}$/s, a $0.1^{\circ}$ error deadband,
and a slew limit of 6 offset units per 10\,ms tick. Full command is reached at
$V/G = 0.375^{\circ}$ of error.

\subsubsection{Insertion law}
Lever deflection commands axial velocity through a deadband:
\begin{equation}
\dot{z} = k_v\,\mathrm{sgn}(s)\,\frac{\max(0,\,|s| - 5^{\circ})}{22^{\circ}}
\end{equation}
capped at 1.5\,mm/s, with the commanded depth bounded to $-12$/$+15$\,mm about
the home position.

\subsection{Timing and Response}
End-to-end latency from handle motion to robot motion was measured optically
with a 240\,Hz camera over 50 trials, counting frames from input onset to first
robot movement: $<10$\,ms.

\subsection{Feedback}
The gimbal motors provide DOF locking, return-to-center, and resistive workspace
boundaries, and a stylus vibrotactile unit signals proximity events.

Force rendering was disabled for this study, with the drive stage
disconnected throughout, and no haptic condition is evaluated here.
The only feedback active during the trials was the vibrotactile workspace-edge
cue, a repeating pulse fired while a commanded robot axis was within
$0.5^{\circ}$ of its $\pm 8^{\circ}$ limit, corresponding to $\pm 22.5^{\circ}$
of handle deflection.

\subsection{Workspace}
The grip point traverses a spherical cap of radius 100\,mm, $\pm 40^{\circ}$ per
axis. With motion scaling factor $g = 3.0$, the instrument sweeps $\pm 8^{\circ}$ about the trocar.

\subsection{Software and Logging}
The host reads the input device at $\sim$490\,Hz, resolves the multiplexer, applies
the frame rotation and scaling, evaluates the control law, and issues joint
commands to the robot controller over a direct serial link at 100\,Hz.

Two synchronous logs share a single clock: a kinematics stream at the uplink
rate (raw $\varphi$, $\psi$, $s$, button state) and a control stream at 100\,Hz
(tracking error, per-joint offsets, measured robot angles, depth state, link
status). Each session archives its configuration, so the parameter set under
which any trial was recorded is recoverable.

\section{Methods}
\label{sec:methods}

\subsection{Participants}
Sixteen participants took part: three vitreoretinal surgeons, six ophthalmology
residents, and seven novices without surgical training. The study was
approved by the Medical Ethics Committee of the Technical University of Munich
(2026-396-S-KK) and all participants gave written informed consent.

Experience level characterizes the sample and is not a factor in the analysis;
the relationship between experience and movement is the subject of future work.

\subsection{Task}
Participants performed the navigation training module of an Eyesi Surgical
simulator (Haag-Streit Simulation). Twelve targets are distributed across the
retinal surface and the instrument tip must be brought to each in turn;
participants were required to acquire five. The instruction was to acquire the
targets without injuring the retina. No instruction was given regarding speed
and no time limit was imposed.

\begin{figure}[!t]
\centering
\begin{minipage}[b]{0.32\linewidth}
  \centering
  \includegraphics[width=\linewidth]{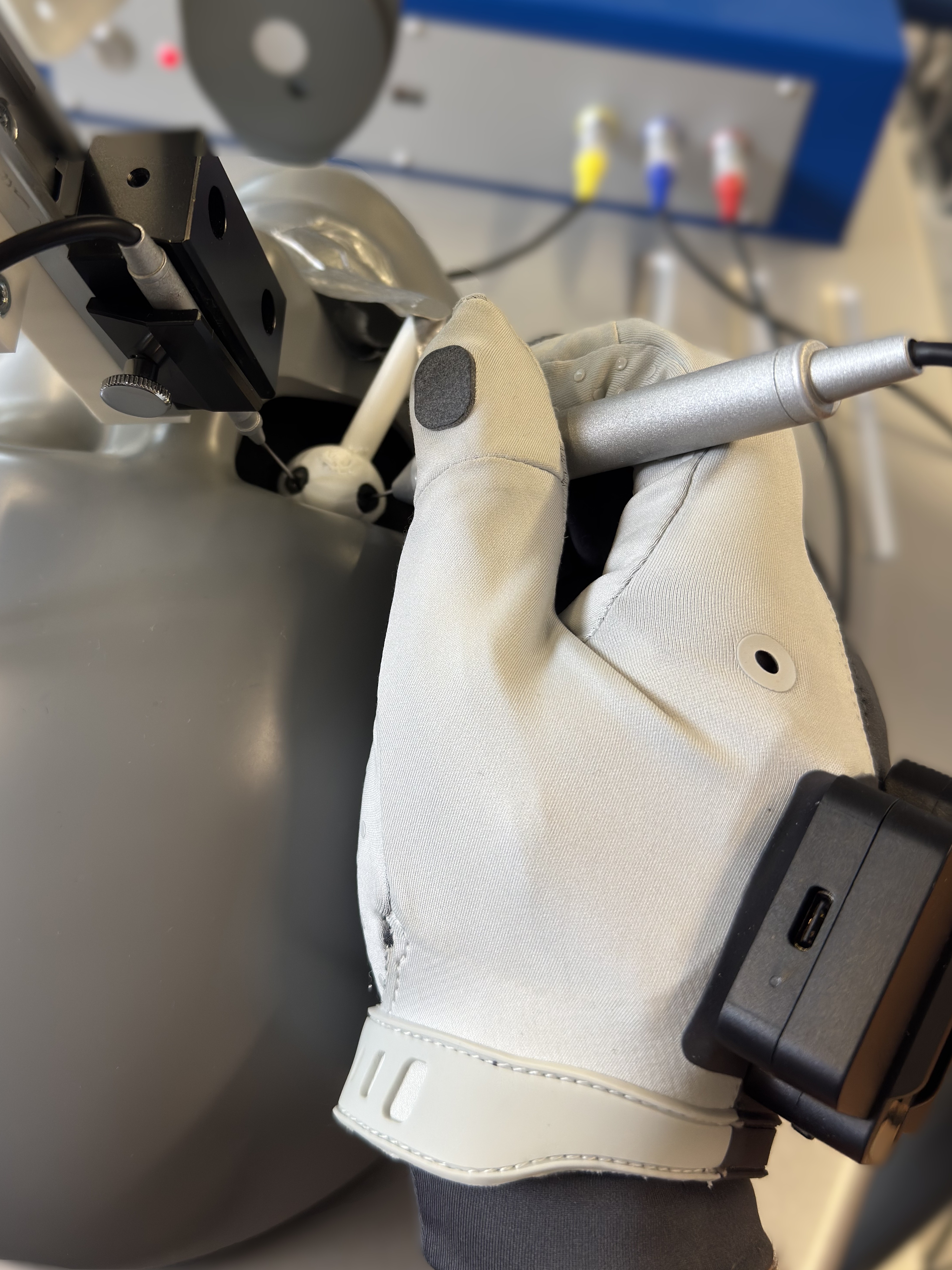}\\[2pt]
  (a)
\end{minipage}
\hfill
\begin{minipage}[b]{0.32\linewidth}
  \centering
  \includegraphics[width=\linewidth]{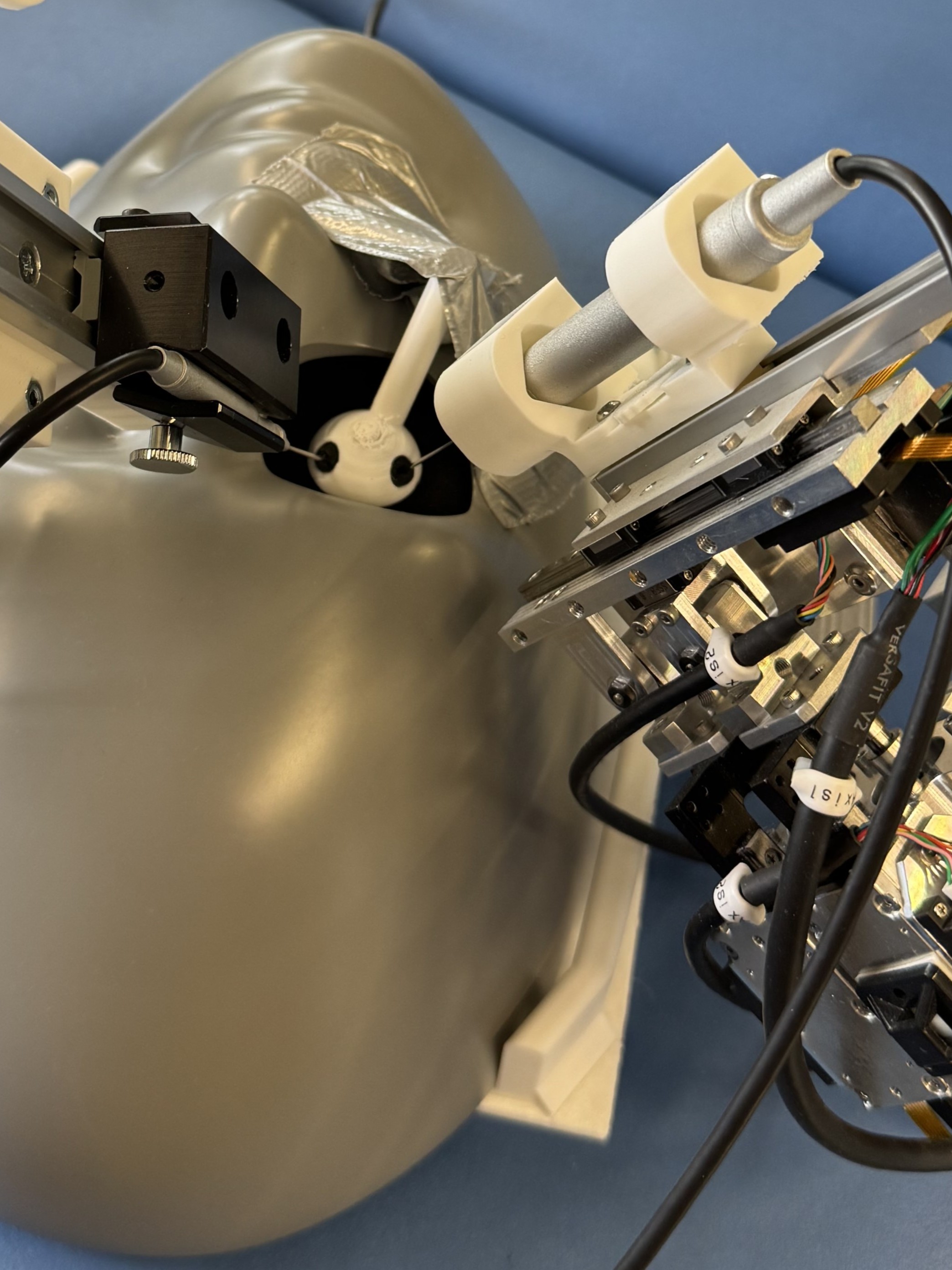}\\[2pt]
  (b)
\end{minipage}
\hfill
\begin{minipage}[b]{0.32\linewidth}
  \centering
  \includegraphics[width=\linewidth]{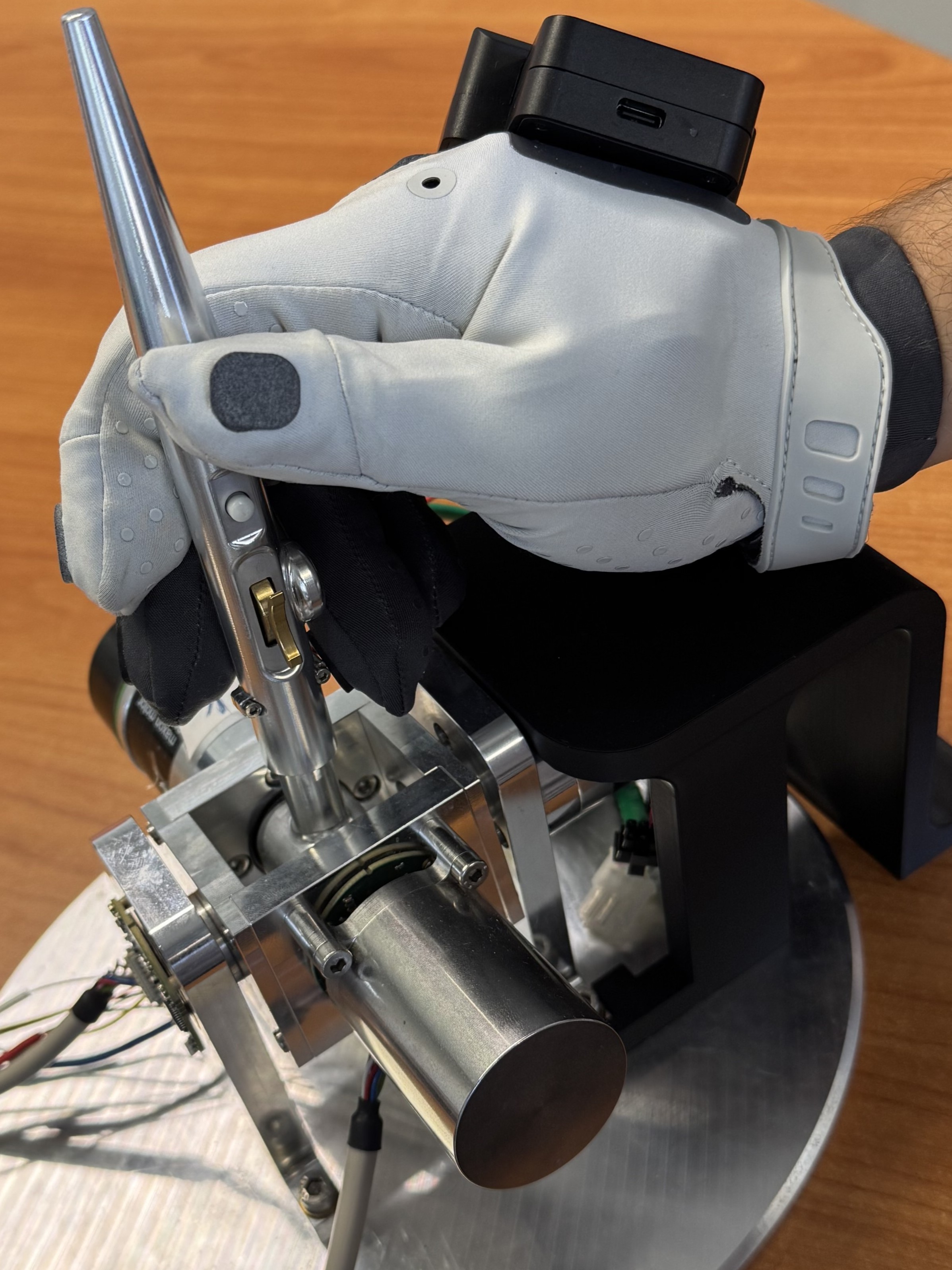}\\[2pt]
  (c)
\end{minipage}
\caption{Experimental setup. (a) Manual condition: the participant holds the
simulator instrument directly and inserts it through the trocar port in the eye
model. (b) Teleoperated condition: the robot holds the same unmodified
instrument, through the same port in the same eye model, and the participant
does not touch it. (c) The EMIRAS input device, held in a pen grasp with the
wrist rested, which commands the robot shown in (b). Between conditions only the
mask and instrument were repositioned, to face either the participant or the
robot; the eye model, the trocar port, the instrument and the light pipe were
common to both. The instrumented glove visible in (a) and (c) was worn only in
the terminal trial of each condition.}
\label{fig:setup}
\end{figure}

\subsection{Conditions and Protocol}
Each participant performed the task manually, holding the simulator instrument
directly, and teleoperated, operating the input device of
Section~\ref{sec:system} which commands the robot holding the same instrument.
Condition order was randomized and a warm-up preceded each condition.

The physical setup was identical across conditions (Fig.~\ref{fig:setup}): the same eye model and mask,
the same trocar port, the same instrument, and a light pipe held fixed and not
manipulated by the participant. Only the mask and instrument were repositioned
between conditions, to face either the participant or the robot. Participants
chose at the outset between the simulator's binocular viewer and its
two-dimensional display and kept that choice throughout, so viewing mode is
constant within participant.

The robot holds the same unmodified instrument the participant holds manually,
through the same port, tracked by the same system, so the control interface is
the only factor that differs. Prior comparisons cannot hold the instrument fixed,
because a robotic instrument is not the tool used manually
(Section~\ref{sec:related}).

Within each condition, participants completed four recorded trials, then one
further trial wearing instrumented gloves (Manus Metagloves). The glove trial
was last in every session because donning and calibrating the gloves interrupts
the session.

\subsection{Recorded Data}
The simulator exports an instrument trajectory at 30\,Hz and a per-trial scoring
file containing targets acquired, retinal and lens injury area, instrument
slip-outs, and an internally computed instrument
path length (the odometer). Tip position is reported with a stated precision of
20--50\,$\mu$m.

In teleoperated trials the host logs the input device at $\sim$490\,Hz
($\varphi$, $\psi$, lever deflection, button state) and the control state at
100\,Hz (tracking error, joint commands, measured robot angles). These two share
a clock. The gloves record 20 finger joint angles at 60\,Hz.

\subsection{Measures}
\label{subsec:measures}
All tool-tip measures are computed in a common frame obtained as follows. The
third column of the simulator's rotation matrix gives the instrument axis in the
eye-fixed frame, so each frame defines a line through the tip. The trocar is the
least-squares intersection of these lines over a trial \cite{b43}; residuals are
at the numerical precision of the data, confirming that the simulator enforces
the remote center of motion exactly. Motion is then expressed in spherical
coordinates about the trocar, giving an insertion depth $L$ and two aiming
angles. This applies identically to both conditions, which matters because
manual surgery has no mode structure on which to segment.

Each trial is bounded by instrument motion rather than by insertion, since in
teleoperated trials the instrument is often inserted before the participant
begins. Each trial's threshold is set relative to that trial's own speed distribution, at 25\% of its 90th-percentile speed. A single absolute threshold would sit above much of the teleoperated speed range while falling below most of the manual range, so it would trim more time from the slower condition and understate the duration difference.

The reported measures are:

\begin{itemize}
\item \emph{Completion time}: duration of the motion-bounded window.
\item \emph{Instrument path length}: total distance travelled by the tip, taken
from the simulator odometer for reasons given in
Section~\ref{subsec:validity}.
\item \emph{Median tip speed}: median instantaneous speed within the window.
\item \emph{Angular range}: the largest angle between any two instrument
orientations in a trial, measured about the trocar.
\item \emph{Movement segments}: contiguous intervals in which tip speed exceeds
5\% of the trial's 90th-percentile speed for at least 0.4\,s. The count measures
how far the task is broken into separate movements.
\item \emph{Clutching measures} (teleoperated only): the aiming duty cycle is
the fraction of the session with the button pressed; engagements are the number
of press--release cycles; the insertion-active fraction is the proportion of the
session with lever deflection beyond its deadband.
\item \emph{Finger measures} (glove trials): total activity is the sum of
standard deviations across the 20 joint angles; pinch variability is the mean
standard deviation of the four pinch measures; submovements are peaks in the
joint-velocity magnitude.
\end{itemize}

\subsection{Analysis}
The teleoperated condition scales angles by $g = 3$, so raw amplitudes differ by
construction and comparisons of absolute excursion would only re-measure the
scaling. Completion time, path length, rates and counts are unaffected by this
and are compared directly.

All comparisons use linear mixed models across the four ungloved trials, with
condition, trial index and their interaction as fixed effects and a
per-participant random intercept. Trial index is centered at the fourth trial, so
the condition coefficient is the difference at the most practised trial and the
interaction shows how that difference changes with practice. Descriptive values
are medians with interquartile ranges. Glove trials, of which there is one per
condition, are compared with the Wilcoxon signed-rank test.

Given sixteen participants, effect sizes and confidence intervals rather than
significance thresholds are the primary evidence, and non-significant
differences are described as such rather than as trends.

\section{Results}
\label{sec:results}
Of 160 scheduled sessions, six trajectory exports were returned empty owing to a
fault at the export stage and three sessions were not recorded. Two further
trials were excluded as aborted recordings, on a criterion independent of task
outcome: fewer than 10\,s of instrument data, or an empty export that could not
be verified. \textbf{149 trials were analyzed}, of which 122 ungloved trials
enter the mixed models and 27 glove trials carry tool-tip data.

\subsection{Task Outcome}
\label{subsec:outcome}
Every participant acquired all five targets in both conditions, and no retinal or
lens injury occurred in any trial of either condition. Across the study there was
one task failure: one participant acquired no targets in a teleoperated glove
trial after 114.6\,s, having acquired all five manually.

The outcome measure is at ceiling by design. Participants were asked for five of
twelve targets, so completion cannot discriminate once both conditions clear the
requirement. What this establishes is that the interface was sufficient for the
task as specified, not that performance was equally precise.

Instrument slip-outs were fewer under teleoperation (median 0.5 against 2.0) and
the simulator's composite score was slightly lower (24 against 26); neither
difference is significant.

To ask a more discriminating question than completion, we examined how each
approach terminated, using the movement segmentation of
Section~\ref{subsec:measures}. Terminal approach speed, as a fraction of each
segment's peak, was lower under teleoperation (0.19 against 0.23, $p = 0.007$),
which is what a rate ceiling produces. Endpoint direction reversals and dwell
near rest did not differ. The measured terminal-approach variables did not reveal a difference consistent with poorer endpoint control under teleoperation. However, the study was not designed or powered to test precision equivalence, and these findings should therefore not be interpreted as demonstrating equivalent precision between conditions.

\subsection{Measurement Validity}
\label{subsec:validity}
Path length is a sum of absolute displacements, so position noise adds to it
rather than averaging out, in proportion to the number of samples. Teleoperated
trials run three times longer at a quarter of the speed, so any noise floor
inflates the reconstruction more in that condition.

\begin{figure}[!t]
\centering
\includegraphics[width=0.85\linewidth]{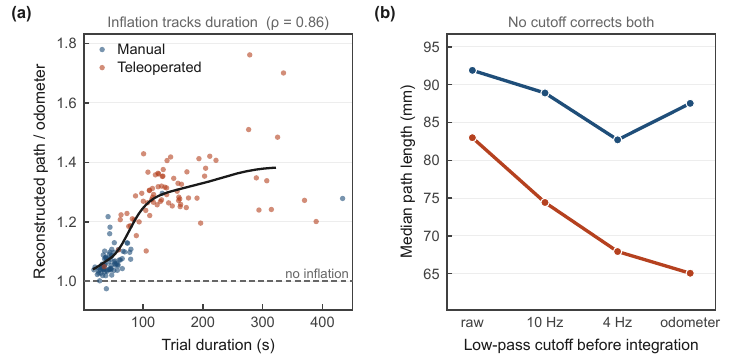}
\caption{Path-length reconstruction error. Left: reconstructed path divided by
the simulator odometer, against trial duration. Right: condition medians at
successive low-pass cutoffs, with the odometer for reference.}
\label{fig:inflation}
\end{figure}

Reconstructing path from the 30\,Hz export and comparing it against the
simulator's internal odometer, which is computed within the simulation at full
rate, quantifies the error (Table~\ref{tab:noise} and Fig.~\ref{fig:inflation}). The reconstruction
over-reads by 5\% in manual trials and 29\% in teleoperated ones, and the
inflation is strongly rank-correlated with trial duration (Spearman's $\rho = 0.86$), which is what accumulating noise predicts. The 10th-percentile
inter-sample displacement is 15\,$\mu$m, below the stated 20--50\,$\mu$m
precision, so many samples carry no resolvable motion. Low-pass filtering does
not fix this: at a 4\,Hz cutoff the reconstruction under-reads the manual
condition while still over-reading the teleoperated one, because the cutoff that
removes the teleoperated noise floor also removes real manual motion.

Reconstructed path length is therefore not comparable between
conditions, and all path results use the simulator odometer. The same hazard
applies to any study comparing conditions of unequal duration using a decimated
position export, and is invisible without an independent path measure.

\begin{table}[!t]
\caption{\textbf{Reconstructed Path Against Simulator Odometer}}
\label{tab:noise}
\renewcommand{\arraystretch}{1.25}
\centering
\small
\begin{tabular}{@{}lrrrrr@{}}
\toprule
& \multicolumn{4}{c}{Path length  (median, mm)} & \\
\cmidrule(lr){2-5}
Condition & Odometer & Raw & 10\,Hz & 4\,Hz & Inflation \\
\midrule
Manual       & 87.5 & 91.9 & 88.9 & 82.7 & 1.05$\times$ \\
Teleoperated & 64.8 & 83.6 & 73.9 & 67.9 & 1.29$\times$ \\
\bottomrule
\end{tabular}
\end{table}

\subsection{How the Task Was Performed}
\label{subsec:main}
Table~\ref{tab:main} reports the comparison across the four ungloved trials. Every measure differs between conditions, and the differences separate into two groups: those where the gap narrows with practice, and those where it does not detectably change. The distinction turns out to track what sets each measure, and Section~\ref{sec:discussion} returns to it.

\begin{figure}[!t]
\centering
\includegraphics[width=0.62\linewidth]{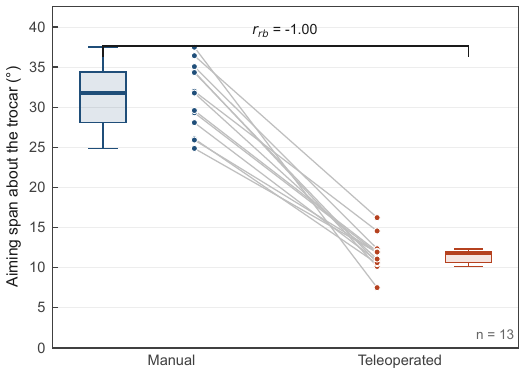}
\caption{Angular range covered at trial 4, per participant, for the 13 of 16 participants with data in both conditions.}
\label{fig:workspace}
\end{figure}

Teleoperated trials took three times as long at roughly a quarter of the median
speed, over an instrument path about a quarter shorter. Operators covered less
than half the angular territory (Fig.~\ref{fig:workspace}), and
the task was broken into 3.5 times as many separate movements.

\begin{table}[!t]
\caption{\textbf{Task Performance. Median (Interquartile Range) at Trial 4, and the Mixed-Model Condition Effect Across Trials 1--4}}
\label{tab:main}
\renewcommand{\arraystretch}{1.25}
\setlength{\tabcolsep}{4pt}
\centering
\small
\begin{tabular}{@{}lrrrlc@{}}
\toprule
&  \\
Measure & Manual & Teleoperated & Effect & [95\% CI] & Practice-dependent \\
\midrule
Completion time (s)        & 37.2 (30.5--45.4)  & 111.0 (88.0--135.6)  & $+71.8$ & [$+46.5$, $+97.1$] & Yes ($-27$\,s/trial) \\
Instrument path (mm)       & 79.3 (74.5--118.3) & 62.9 (58.2--71.0)    & $-25.8$ & [$-40.0$, $-11.6$] & Not detectably \\
Median tip speed (mm/s)    & 1.2 (1.0--1.4)     & 0.3 (0.3--0.3)       & $-0.9$ & [$-1.0$, $-0.8$]    & Not detectably \\
Angular range ($^{\circ}$) & 30.7 (26.6--34.4)  & 11.5 (10.6--12.0)    & $-17.1$ & [$-19.9$, $-14.4$] & Not detectably \\
Movement segments          & 16.5 (11.5--18.0)  & 58.0 (45.0--73.5)    & $+41.6$ & [$+26.1$, $+57.0$] & Yes ($-14$/trial) \\
\bottomrule
\end{tabular}
\end{table}

\subsection{Rate Ceilings}
The robot's rotational and translational joints are commanded at up to
$1.5^{\circ}$/s and 1.5\,mm/s. This is not a conservative safety margin but the
operating limit of the present manipulator: above it, the joints do not hold the
positioning accuracy the task requires. Manual peak aiming reached $14.4^{\circ}$/s, roughly ten times that limit, and the tracking law was saturated on
a median 6\% and 12\% of control ticks across the two rotational axes.

The observed duration difference is quantitatively consistent with the combination of shorter teleoperated path length and substantially lower movement speed. A path ratio of 0.79 and a mean-speed ratio of 0.25 predict a duration ratio of 3.19, compared with 2.98 observed. Together with the frequent
saturation of the tracking law, this indicates that the robot's configured velocity limit is a major contributor to the temporal cost of teleoperation.

\subsection{Clutching}
\label{subsec:clutch}
Rotational commands reached the robot on 38\% of the session
(Table~\ref{tab:control}). Participants made a median of 29 press--release cycles
per trial, engaging for 0.7\,s and releasing for 1.6\,s at a time.

The mode design accounts for part of this. The button is a multiplexer:
releasing it is the only way to command insertion, so any alternation between
aiming and insertion forces a release. But five targets imply five to ten forced
cycles against the 29 observed, so most engagements are not explained by mode
necessity. The remainder may be fine repositioning, hand comfort or hesitation.

The lever was active during 22\% of trial time, compared with an aiming duty cycle of 38\%. Thus, 40\% of trial time contained neither an active aiming command nor an active insertion command. This interval may include visual assessment, preparation for mode transitions, fine repositioning of the hand, or other operator-level control behavior. The present data do not distinguish among these possibilities.

\begin{table}[!t]
\caption{\textbf{Clutching in Teleoperated Trials Across All 16 Participants}}
\label{tab:control}
\renewcommand{\arraystretch}{1.25}
\centering
\small
\begin{tabular}{@{}lr@{}}
\toprule
Measure & Median (IQR) \\
\midrule
Aiming duty cycle          & 0.38 (0.34--0.44) \\
Engagements per trial      & 29 (18--40) \\
Engagement duration (s)    & 0.65 (0.56--1.28) \\
Release duration (s)       & 1.6 (1.1--1.8) \\
Insertion-active fraction  & 0.22 (0.20--0.26) \\
\bottomrule
\end{tabular}
\end{table}

\subsection{Practice}
\begin{figure}[!t]
\centering
\includegraphics[width=\linewidth]{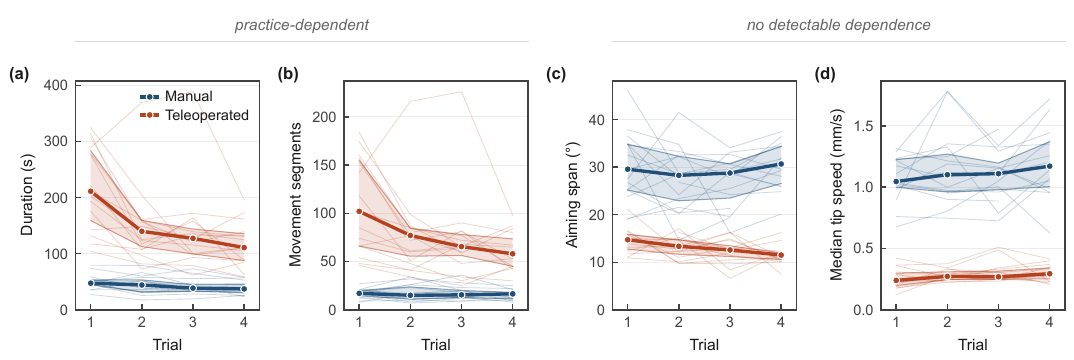}
\caption{Practice across trials 1--4. Faint lines are participants, bold lines
condition medians. The conditions converge in the two left panels and remain separated by a constant margin in the two right panels.}
\label{fig:learning}
\end{figure}

Two measures improved substantially across the four trials: completion time by
27\,s per trial and movement segments by 14. Neither had plateaued by the fourth
trial. The teleoperated deficits reported above are
therefore upper bounds on a still-improving skill (Fig.~\ref{fig:learning}).

The remaining measures showed no detectable change in the gap between
conditions. These are the measures governed by the robot's velocity limit and joint range,
which is what a fixed hardware constraint predicts. A non-significant interaction
is not evidence that a gap is constant, and for instrument path in particular the
interaction interval is wide relative to the effect, so that classification is
uncertain.

Separating participants by experience refines this. The nine surgeons and
residents showed a flat manual trend, consistent with working at asymptote on a
familiar tool; the seven novices improved manually as well, which is expected
since they were naive to both. Teleoperated improvement was steep in both
subgroups, so the observation does not depend on the split.

\subsection{Hand Kinematics}
\label{subsec:hand}
In the glove trials, finger activity approximately doubled under teleoperation
(134.9 against 71.2 summed joint standard deviations, $p = 0.001$), pinch
variability tripled (10.0 against 3.3, $p = 0.0002$), and finger submovements
doubled (115 against 55, $p = 0.0046$). Every participant moved in the same
direction on the first two measures.

Absolute tremor-band power (8--12\,Hz) at the fingers did not fall: it increased
in 10 of 13 participants, though not significantly ($p = 0.17$). The
teleoperated hand is therefore not steadier. It is busier.

What reaches the instrument is a separate matter, and it follows from the
scaling rather than from the hand. Handle deflection of $0.0131^{\circ}$ RMS
corresponds to 23\,$\mu$m RMS at the 100\,mm grip radius; dividing by $g = 3$ and
projecting through the working depth gives 1.55\,$\mu$m RMS at the tip, against
38\,$\mu$m RMS reported for unassisted vitreoretinal surgery \cite{b1}. This is a
projection, not a measurement: the 30\,Hz export cannot resolve the 8--12\,Hz
band at the tip, and most of the reduction is arithmetic, following from $g$ and
the grip-to-depth ratio.

Three caveats bound the finger results. The glove trial was last in every
session, so the effect is confounded with fatigue and order; gloved trials were
also slower than ungloved ones in the manual condition (48.5 against 37.2\,s,
$p = 0.003$), so the gloves themselves perturbed performance. It is a single
trial per condition. And the gloves are a commercial motion-capture product
whose frequency response in the tremor band is uncharacterized, so the tremor
comparison is relative rather than calibrated.

\section{Discussion}
\label{sec:discussion}

\subsection{What the Interface Preserves and What It Changes}
Task success and movement execution diverged. Both interfaces satisfied the
predefined task-success criterion, but teleoperation substantially altered how
that success was achieved. Movements were slower, more fragmented, and confined
to a smaller angular range. This distinction is important because the device
was explicitly designed around manual-like posture and aiming gestures. The
results show that reproducing the physical geometry of manual manipulation at
the input device does not necessarily preserve manual motor behavior once
motion scaling, mode switching, workspace constraints, and robot dynamics
intervene.

The two modes differ in how far they preserve manual technique, deliberately.
Aiming is position-mapped with a fixed 3:1 angular reduction. Insertion is
rate-controlled through a deadbanded lever: a surgeon inserting an instrument
moves the hand a distance proportional to the insertion, whereas here lever
deflection is proportional to insertion \emph{speed}. The design premise holds
for aiming and is abandoned for insertion.

\subsection{The Velocity Limit Is a Major Constraint}
Manual peak aiming exceeded the robot's limit by roughly tenfold, and
completion time follows from the speed reduction with no residual. The threefold difference in completion time is therefore close to entailed
rather than surprising: it is what a manipulator limited to a quarter of manual mean velocity must produce over a comparable path. What the measurement adds is
the magnitude, and the observation that the difference composes exactly from
path and speed with no residual, so nothing else contributes.

The measures governed by the velocity limit and the joint range are also those showing the smallest practice effects, while the measures reflecting how the
operator organizes the task improved substantially.

\subsection{Clutching}
Operators engaged the rotational command 29 times per trial, and the mode design
explains perhaps a third of that. The multiplexer forces a release whenever
aiming alternates with insertion, but not thirty times for five targets.

This matters for redesign. Allowing insertion without releasing the aiming
command, or switching modes automatically on depth, would remove the forced
component. How large that component is remains open, and the answer determines
how much such a change is worth: if most releases are followed by insertion the
multiplexer is the right target, and if they are not, the target is whatever
makes operators release without inserting.

\subsection{The Hand Works Harder}
Reducing instrument degrees of freedom did not reduce manual work. It
redistributed it, from continuous guidance of a tool into repeated press-release
cycles, mode selection and lever modulation. Finger activity doubled and pinch
variability tripled.

This complicates any ergonomic argument for reducing degrees of freedom, and it
points at the same target as the previous subsection: effort is concentrated in
the engagement cycle. Finger-level measurement in ophthalmic microsurgery is
recent and confined to manual technique \cite{b40}, so there is little basis for
calibrating the magnitude. The direction is robust; the magnitude is not.

\subsection{Design Implications}
Two follow from the measurements.

First, the temporal cost is bounded by the robot's velocity limit, which is set
by the accuracy the joints can hold rather than by the control software. Reducing
it requires a manipulator that stays accurate at higher speed, not a parameter
change. A limit that varied with context could help: it binds only during fast
excursions, while median tracking error stayed at $0.083^{\circ}$, and a speed
constraint is most needed near tissue and least during gross repositioning.

Second, the multiplexed button is worth reconsidering. Roughly 40\% of the session is spent
in neither aiming nor inserting, and the engagement cycle is what loads the hand.

\subsection{Limitations}
The study was conducted in a simulator. Trocar friction, tissue compliance and
the consequences of error are absent, and the remote center of motion is enforced
exactly, so the trocar-recovery method was validated under ideal conditions.

A single navigation task was used, exercising aiming and insertion but not
membrane peeling, injection or bimanual work, and the light pipe was fixed.

Force rendering was disabled throughout, so this work characterizes a transparent
input device and says nothing about its haptic capability.

No subjective workload instrument was used. For a study reporting that the hand
works harder, the absence of a perceived-effort measure is a real gap: the finger
kinematics establish that more motion occurs, not that operators experience more
effort.

The instrumented gloves perturbed performance and were worn only in a terminal
trial, confounding their effect with fatigue and order. No tip-tremor measurement
is available in either condition.

\section{Conclusion}
\label{sec:conclusion}
We compared manual and teleoperated intraocular instrument motion with the trocar
constraint, the instrument, the eye model and the tracking source common to both
conditions, so that the control interface was the only factor varied.

Task outcome was equal, though the measure was at ceiling by design. Execution differed on every measure:
teleoperated trials took three times as long at a quarter of the speed, covered
under 40\% of the angular territory, and were broken into 3.5 times as many movements.
Completion time and fragmentation improved substantially across four trials and
had not plateaued; the measures set by the robot's rate limit did not.

The temporal cost follows from the robot's velocity limit, which is set by the
accuracy its joints hold and is therefore a property of the manipulator rather
than of the control software. The engagement frequency and the lever deadband are
not: both follow from the control design and could be changed without new
hardware.

One finding runs against the design's expectations. Reducing instrument degrees
of freedom did not reduce manual effort: finger activity doubled and pinch
variability tripled as work moved into repeated press-release cycles.

Two contributions generalize beyond this device. Recovering the remote center of
motion from an instrument trajectory and decomposing motion about it requires
only a tracked instrument and applies identically to manual and teleoperated
operation, which is what makes the comparison possible. And reconstructing path
length from a decimated position export inflates it in proportion to trial
duration, which here nearly produced a false equivalence between conditions of
unequal length. Both are simple to apply, and we would argue that any input device described as preserving natural technique should be measured this way rather than asserted to do so.

\section*{Conflict of Interest}
The input device evaluated here was designed and built by the authors, and
the robot it commands was developed and previously characterized by the same
group \cite{b3}. This is therefore an evaluation of an in-house device by its
developers. This work was partially supported by NSK Ltd.; the funder had no
role in the study design, data collection, analysis, interpretation, or the
decision to submit for publication. The authors declare no other competing
interests.

\section*{Data and Code Availability}
The analysis pipeline (trajectory parsing, remote-center recovery, the depth and aiming decomposition, and every derived measure reported here) and participant data are available from the
corresponding author subject to the terms of the ethics approval.

\section*{Acknowledgment}
The authors thank Lennart Frese and Andrea Ross for assistance with data collection, and Haag-Streit
Simulation for providing trajectory and scoring export support. During the preparation of this manuscript, the authors used
Claude (Anthropic) to assist with language editing, LaTeX formatting, and drafting of figure captions. All content was subsequently reviewed, verified and edited by the authors, who take full responsibility for the accuracy and integrity of the work.

\section*{Author Biographies}

\noindent\textbf{Korab Hoxha} received his B.Sc. in Mechanical Engineering with a focus on Medical Technology from the Technical University of Munich (TUM) in 2020, which included an ERASMUS+ semester at Aston University. He obtained his M.Sc. in Medical Technology from TUM in 2023. He is currently pursuing his Ph.D. as a Research Associate at TUM Klinikum rechts der Isar, since May 2023, where he is developing a robotic system for eye surgery. His research interests include surgical robotics, biomechanics, and automation in medicine.

\medskip
\noindent\textbf{Mirza Imamovic} earned his bachelor's degree in Mechanical Engineering from the University of Tuzla and his master's degree in Mechatronics and Robotics from the Technical University of Munich. Since 2025, he has been working as a research assistant at the Chair of Ergonomics in collaboration with the Munich Institute of Robotics and Machine Intelligence (MIRMI). He is responsible for establishing and coordinating the Ergonomics Evaluation Lab (EEL). His research focuses on the field of robot-assisted surgery, particularly on the study of teleoperation, haptic feedback, and human-robot interaction in surgical applications.

\medskip
\noindent\textbf{Angelo Henriques} is a Research Associate at the TUM University Hospital within the Medical Autonomy and Precision Surgery (MAPS) group. He is a doctoral candidate at the TUM School of Computation, Information and Technology. His research focuses on integrating scene graphs and multimodal sensor data into robotic surgical workflows to enhance efficiency in ophthalmic microsurgery. He holds both an M.Sc. in Medical Technology and a B.Sc. in Mechanical Engineering from the Technical University of Munich. His interests lie in surgical data science, computer vision, and explainable AI.

\medskip
\noindent\textbf{M. Ali Nasseri} is Professor of Surgical Robotics at TUM and Founding Director of the Medical Autonomy and Precision Surgery (MAPS) Laboratory at TUM University Hospital. He received his Ph.D. in Surgical Robotics from TUM through an interdisciplinary program spanning medicine, engineering, and informatics. He has authored more than 150 scientific publications, patents, and book chapters. His research focuses on surgical robotics, autonomous surgery, surgical intelligence, and AI-driven precision surgery.

\end{document}